\pdfoutput=1

\documentclass[11pt]{article}

\usepackage[]{acl}

\usepackage{times}
\usepackage{latexsym}

\usepackage[T1]{fontenc}

\usepackage[utf8]{inputenc}

\usepackage{microtype}

\usepackage{inconsolata}

\usepackage{amsmath}
\usepackage{amssymb}
\usepackage{mathtools}
\usepackage{amsthm}
\usepackage{url}
\usepackage{amsfonts}       
\usepackage{nicefrac}       
\usepackage{xcolor}         
\usepackage{epsfig}
\usepackage{pifont}
\usepackage{kotex}
\usepackage{enumitem}
\usepackage{verbatim}
\usepackage{bm}
\usepackage{subcaption}
\usepackage{makecell,multirow, tabularx}
\usepackage{algorithm}
\usepackage{url}
\usepackage{balance}
\usepackage{upquote}
\usepackage{appendix}
\usepackage{nicefrac}
\usepackage{physics}
\usepackage{stmaryrd}
\usepackage{pifont}
\usepackage{wrapfig} 

\usepackage{caption}
\usepackage{subcaption}
\usepackage{xcolor}
\usepackage{xspace}
\newcommand\eg{\textit{e.g.}\xspace}

\usepackage{booktabs} 
\usepackage{multirow} 
\usepackage{lipsum} 

\usepackage{algpseudocode}

\title{Interactive Text-to-Image Retrieval with Large Language Models:\\A Plug-and-Play Approach}

\author{Saehyung Lee$^1$\thanks{Equal Contribution}\hspace{4mm} Sangwon Yu$^1$\footnotemark[1]\hspace{4mm} Junsung Park$^1$\hspace{4mm} Jihun Yi$^1$\hspace{4mm} Sungroh Yoon$^{1,2}$\thanks{Corresponding Author}\\
$^1$Department of Electrical and Computer Engineering, Seoul National University \\
$^2$Interdisciplinary Program in Artificial Intelligence, Seoul National University \\ 
{\tt\footnotesize \{halo8218, dbtkddnjs96, jerryray, t080205, sryoon\}@snu.ac.kr}
}

\begin{document}
\maketitle
\begin{abstract}

In this paper, we primarily address the issue of dialogue-form context query within the interactive text-to-image retrieval task. Our methodology, PlugIR, actively utilizes the general instruction-following capability of LLMs in two ways. First, by reformulating the dialogue-form context,
we eliminate the necessity of fine-tuning a retrieval model on existing visual dialogue data, thereby enabling the use of any arbitrary black-box model.
Second, we construct the LLM questioner to generate non-redundant questions about the attributes of the target image, based on the information of retrieval candidate images in the current context. This approach mitigates the issues of noisiness and redundancy in the generated questions. Beyond our methodology, we propose a novel evaluation metric, Best log Rank Integral (BRI), for a comprehensive assessment of the interactive retrieval system. PlugIR demonstrates superior performance compared to both zero-shot and fine-tuned baselines in various benchmarks. Additionally, the two methodologies comprising PlugIR can be flexibly applied together or separately in various situations. Our codes are available at \url{https://github.com/Saehyung-Lee/PlugIR}.
\end{abstract}

\section{Introduction}

Text-to-image retrieval, a task focused on locating target images in an image database that correspond to an input text query, has seen significant advancements thanks to the development of vision-language multimodal models \cite{CLIP,BLIP}. Conventionally, methods in this domain have adopted a single-turn retrieval approach, reliant on the initial text input, which necessitates comprehensive and detailed descriptions from users.
Recently, \citet{levy2023chatting} have introduced a chat-based image retrieval system utilizing large language models (LLMs) \cite{radford2019language} as questioners to facilitate multi-turn dialogues, enhancing retrieval efficiency and performance even with simplistic initial image descriptions given by users.
However, this chat-based retrieval framework confronts certain limitations, including the requirement of fine-tuning to adeptly encode dialogue-style texts, a process that is both resource-intensive and impractical for scalability. Moreover,
the reliance of the LLM questioner on initial descriptions and dialogue histories, without the ability to view the image candidates,
poses a risk of generating queries about non-existent attributes in the target image, based on the LLM's parametric knowledge. 

\begin{figure*}[!t]
    \centering
    \includegraphics[width=\textwidth]{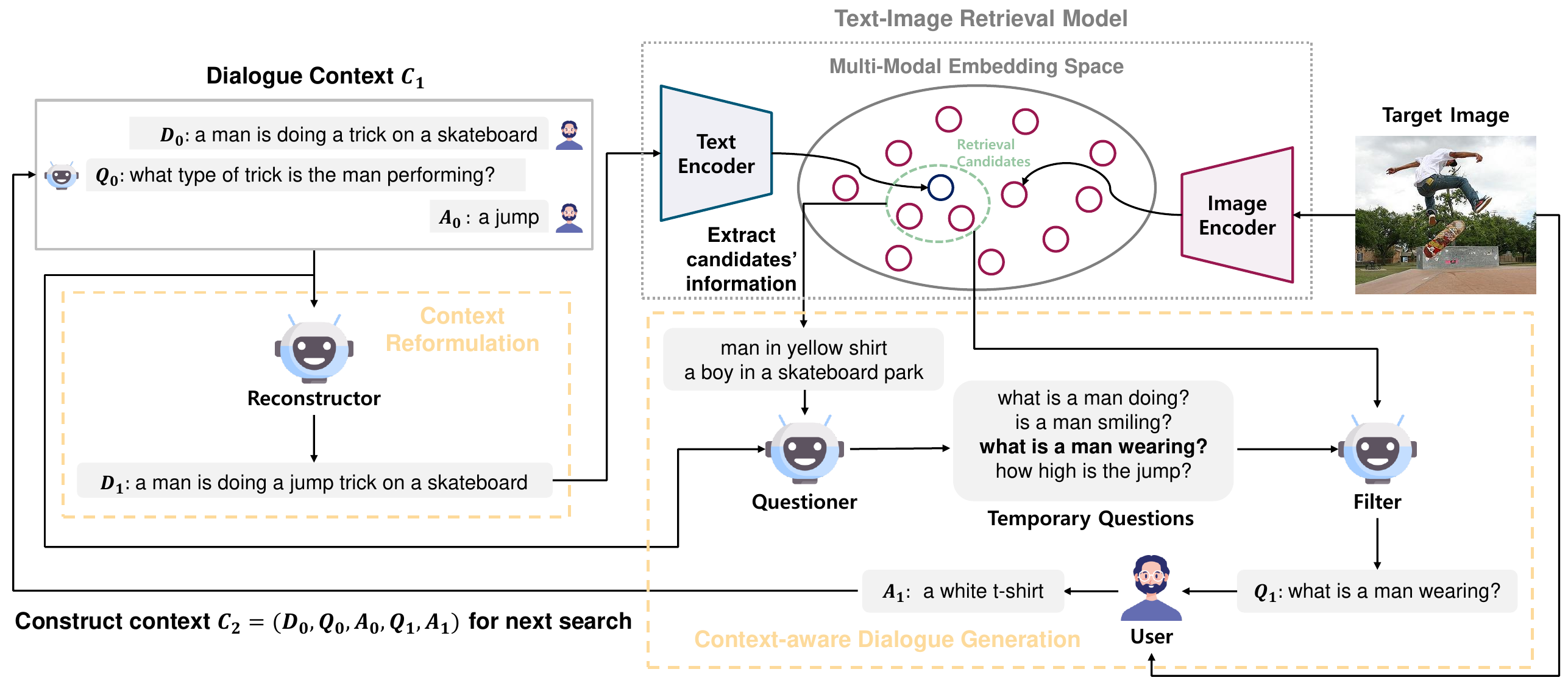} 
    \caption{
    The main framework of the plug-and-play interactive text-to-image retrieval system.
    }
    \label{fig:mainframe}
\end{figure*}
To overcome these challenges, this paper introduces PlugIR, a novel plug-and-play interactive text-to-image retrieval methodology that is tightly coupled with LLMs.
PlugIR comprises two key components: context reformulation and context-aware dialogue generation.
Harnessing the instruction-following proficiency of LLMs, PlugIR reformulates the interaction context between users and questioners into a compatible format for
pre-trained vision-language
models \cite{BLIP}. This process enables the direct application of an array of multimodal retrieval models, including black-box variants, without necessitating further fine-tuning.
Additionally, our approach ensures that the LLM questioner's inquiries are grounded in the context of the retrieval candidates set, thereby allowing it to formulate questions pertinent to the target image's attributes. 
During this process, we inject the retrieval context in text form as an input context for the LLM questioner to reference. Subsequently, our methodology also incorporates a filtering process that selects the most context-aligned, non-repetitive questions, thereby streamlining the search options.
Figure \ref{fig:mainframe} illustrates the overall structure of our proposed interactive text-to-image retrieval system.

We identify three critical aspects for assessing interactive retrieval systems: user satisfaction, efficiency, and ranking improvement significance. We show that
existing metrics, such as Recall@K and Hits@K \cite{recall_loss,levy2023chatting}, fall short in these areas. For instance, Hits@K fails to account for efficiency, which is better with fewer interactions to locate the target image. To resolve these issues, we introduce the \textbf{B}est log \textbf{R}ank \textbf{I}ntegral (BRI) metric. BRI effectively covers all three essential aspects, offering a comprehensive evaluation independent of a specific rank K, unlike Recall@K or Hits@K.
We empirically demonstrate that BRI aligns more closely with human evaluation compared to existing metrics.

Experiments across diverse datasets, including VisDial \cite{VisDial}, COCO \cite{COCO}, and Flickr30k \cite{Flickr30k}, show that PlugIR significantly outperforms the existing
interactive retrieval systems using zero-shot or fine-tuned models \cite{levy2023chatting}.
Moreover, our approach shows significant adaptability when applied to diverse retrieval models, including black-box models.
This compatibility extends the utility of our approach, allowing it to be adapted to a broader spectrum of applications and scenarios. 
We summarize our contributions as follows:
\begin{itemize}[leftmargin=*, nolistsep]
    \item{
    We present the first empirical evidence showing that zero-shot models struggle to understand dialogues and introduce a context reformulation method as a solution. This method does not necessitate fine-tuning the retrieval model.
    }
    \item{We propose a LLM questioner designed to address the searching bottleneck issue caused by noisy and redundant questions}
    \item{
    We introduce BRI, a novel metric aligned with human judgment, specifically designed to enable comprehensive and quantifiable evaluation of interactive retrieval systems.
    }
    \item{
    We verify the effectiveness of our framework across a diverse range of environments, highlighting its versatile plug-and-play capabilities.
    }
\end{itemize}

\section{Related Work}
\textbf{Text-to-Image retrieval task}
The task of retrieving a target image from an image pool through user interaction is known as text-to-image retrieval.
Various methods have been proposed for retrieving target images using various forms of user interaction~\cite{levy2023data,liu2021image,vo2019composing,wu2021fashion}.
Notably, ChatIR~\cite{levy2023chatting} introduced a method for image retrieval through dialogue between a user and an automated system. Further related works are provided in Appendix \ref{related_work} due to the space limitations.

\paragraph{Vision-Language models}

Vision-language models (VLMs) have emerged as a pivotal area in AI research, aiming to bridge the gap between visual and textual understanding. 
CLIP \cite{CLIP} introduces the vision-language landscape by leveraging a contrastive learning framework. 
By jointly embedding images and their associated text descriptions, CLIP demonstrates a robust ability to perform various vision-language tasks, especially in zero-shot classification across a wide range of concepts. 
Subsequently proposed BLIP \cite{BLIP} goes further by introducing a model capable of both understanding and generation task, and addressing the issue of noisy captions in web data used for pre-training. As a result, BLIP exhibits exceptional performance in zero-shot image-text retrieval.
Several pioneering models, such as BLIP-2 \cite{BLIP2}, have significantly advanced this field, exhibiting remarkable capabilities in cross-modal representation learning, and we make use of these VLMs as our text-to-image retrieval models.

\paragraph{Large language models}
Beginning with the Generative Pretrained Transformer (GPT) series~\cite{radford2018improving, radford2019language, openai2023gpt4}, a variety of works have proposed scaling up the parameters of language models to the billion-scale~\cite{touvron2023llama,touvron2023llama2}. 
Increasing the number of parameters has not only enhanced the performance of language models but also revealed various emergent abilities~\cite{wei2022emergent}, which have enabled remarkable performance in a range of downstream tasks, including zero-shot and few-shot learning.
Beyond training high-performance LLMs, topics on techniques like Chain-of-Thought prompting~\cite{wei2022chain} and self-consistency~\cite{wang2022self} to effectively extract answers from trained LLMs are active research area.

\section{Method} \label{method}

\subsection{Preliminaries: Interactive Text-to-Image Retrieval}
Interactive text-to-image retrieval is a multi-turn task that begins with a simple initial description, $D_0$, provided by the user. This task involves a dialogue between the user and the retrieval system about the image corresponding to $D_0$ (the target image), forming a context used as the search query for the target image in each turn (round). In each round $t$, the retrieval system generates a question $Q_t$ about the target image, and the user responds with an answer $A_t$, thereby creating the dialogue context $C_t=(D_0, Q_0, A_0, …, Q_t, A_t)$ for that round. This dialogue context undergoes appropriate processing, such as concatenating all text elements, to form a single text query used for image searching in that round. During the image searching, the retrieval system matches all images in the connected image pool with the text query and ranks them based on a similarity score. The performance of the retrieval system can be evaluated based on the retrieval rank of the target image.

For evaluation, two primary metrics are commonly used: Recall@K and Hits@K.
When evaluated using Recall@K, success is determined if the target image's rank computed in the current round is among the top K. For Hits@K, success is defined as the target image appearing in the top-K results at any round up to the current one.


\subsection{Context Reformulation}\label{sec:cotext_reformulation}
\paragraph{Do zero-shot models understand dialogs?}\label{sec:recon_motivation}
To demonstrate the necessity of the proposed method, we assess the degree to which zero-shot models comprehend and effectively employ given dialogues in the interactive text-to-image retrieval task. We specifically track changes in the retrieval performance of zero-shot models, which comprise three white-box models (CLIP, BLIP, and BLIP-2) and one black-box model\footnote{\url{https://docs.aws.amazon.com/bedrock/latest/userguide/titan-multiemb-models.html}}, by incrementally providing an additional question-answer pair related to the target image over 10 rounds. Thus, in the 10th round, the input query is a dialogue encompassing one image caption and 10 question-answer pairs. We posit that if a zero-shot model is capable of understanding dialogues and utilizing them effectively in the image retrieval task, it will exhibit enhanced performance in the later rounds compared to its initial performance in round 0, which solely involved the use of the image caption.

\begin{figure}[!t]
{
\begin{center}
\centerline{\includegraphics[width=\columnwidth]{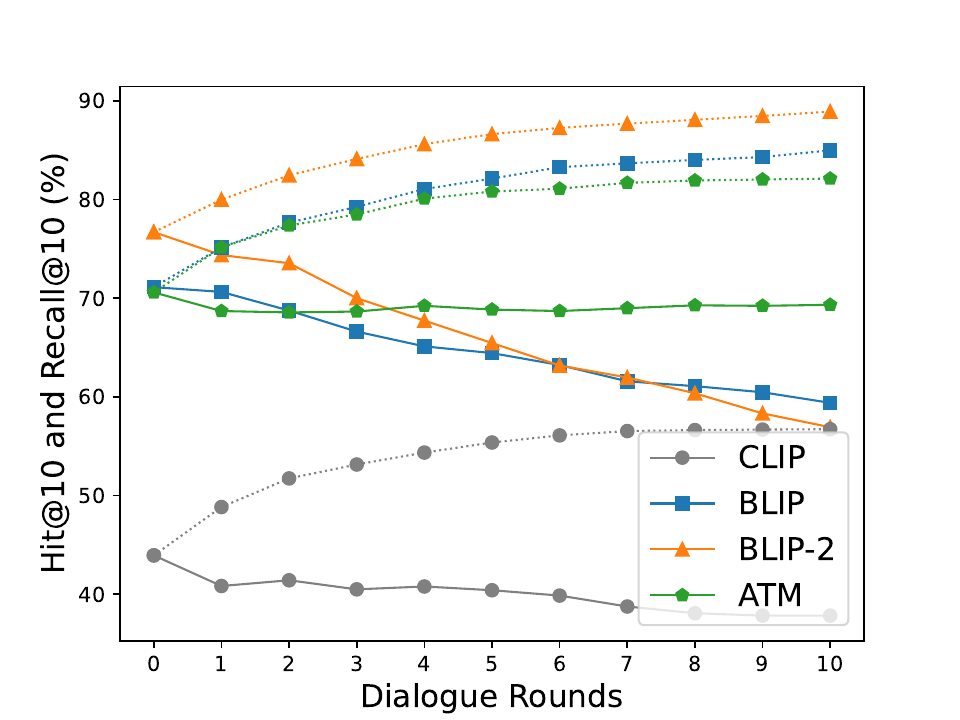}}
\caption{
Round-by-round text-to-image retrieval performances of CLIP, BLIP, BLIP-2, and the Amazon Titan multimodal foundation model (ATM). In the 0th round, an image caption is provided as the query, and with each subsequent round, a single question-answer pair is added. Solid lines represent Recall@10, while dotted lines indicate Hits@10.
}
\label{fig:recon_motivationy}
\end{center}
}
\end{figure}
Figure \ref{fig:recon_motivationy} illustrates a progressive improvement in the Hits@10 scores of all tested zero-shot models across successive rounds. This trend suggests that some query samples, initially unsuccessful in retrieval, achieve success as the dialogues are enriched in later rounds. However, we advise against hastily concluding that dialogues are effective as input queries for zero-shot models based solely on these observations. Our analysis, informed more by the Recall@10 than the Hits@10 scores, leads to a different conclusion: the zero-shot models appear to struggle with comprehending dialogues in the text-to-image retrieval task.

In fact, Hits@K scores can increase over consecutive rounds by simply adding noise to the similarity matrix between image captions and candidate images.
This occurs because Hits@K requires only one successful retrieval attempt at any point up to each round.
In contrast, Recall@K reflects the quantity of information present in ``each'' round's queries in the text-to-image retrieval task.
Figure \ref{fig:recon_motivationy} shows that all the retrieval models under study achieve their highest Recall@10 scores when using only image captions as input queries. Notably, the CLIP, BLIP, and BLIP-2 models experience a decrease in Recall@10 as the round progresses. This trend implies that the appended dialogues, in the context of these zero-shot models, predominantly function as noise. In CLIP, BLIP, and BLIP-2, the effect of noise becomes more pronounced with the increased dialogue length. The Amazon Titan multimodal foundation model (ATM), while not showing a decrease in Recall@10 with longer dialogues, does not exhibit enhanced performance either, suggesting that the added dialogues may not substantially contribute to the informative context.

\paragraph{A plug-and-play approach.}
To overcome the challenge of zero-shot retrieval models not effectively using dialogues in text-to-image retrieval tasks, one strategy could be the fine-tuning of pre-trained retrieval models using datasets that consist of image and dialogue pairs. For instance, \citet{levy2023chatting} has fine-tuned the BLIP model on VisDial to attain higher Hits@K scores. We provide empirical evidence in Section \ref{sec:dialogues_utilization}, illustrating that this method can equip retrieval models with the capability to comprehend dialogues. However, the implementation of such tuning-based approaches depends on certain conditions that are not always feasible: (\romannumeral 1) Access must be available to the retrieval model parameters; (\romannumeral 2) Sufficient and suitable training data must be obtained. For example, this method is not applicable to black-box retrieval models like ATM.


In this study, we investigate a novel approach that adapts text queries to be better understood by retrieval models, rather than modifying the retrieval models to accommodate the format of the text queries. More specifically, rather than directly using dialogues as input queries, we utilize LLMs to convert dialogues into a format (\eg, caption-style) more aligned with the training data distribution of the retrieval models. This strategy effectively bypasses the constraints associated with tuning-based methods, as it does not necessitate the fine-tuning of the retrieval models.
The text prompts used for the context reformulation can be found in Appendix \ref{prompt}.
\subsection{Context-aware Dialogue Generation}
\paragraph{Is the additional information in dialogues\\ actually effective?}
The motivation for the reformulation proposed in the previous section is based on the observation that the dialogue form tends to function more as noise than as useful information for a pre-trained retriever. In this section, we aim to delve beyond the form of the context and focus on the actual contents of the context. We identify two key issues when relying solely on the dialogue context to generate questions about the target image. Firstly, the generated questions may inquire about attributes that are unrelated to the target image. For example, questions asking the objects not in the target image are likely to elicit negative responses. This case itself may function as noise within the dialogue context. Consequently, compared to previous rounds, the context representation introduces more confusion in the retrieval process, leading to a decrease in retrieval performance.

The second is the potential redundancy of generated questions. In the question generation process, general questions like ``What is the person in the photo doing?'' can often be answered based on the information already available in the dialogue context,
without needing to view the target image.
In such cases, the question-answer pair
also fails to provide valuable additional information, resulting in redundancy. Consequently, this redundancy does not contribute to enhancing retrieval performance in subsequent rounds.
In the following sections, we address these issues and propose a questioner structure that can be flexibly applied in various situations, effectively tackling the challenges of noise and redundancy in dialogues.

\begin{algorithm}[t]
\caption{Retrieval Context Extraction}
\label{alg:extract}
\begin{algorithmic}[1]
\State \textbf{Input:} dialogue context $\mathbf{c}$, image pool $\mathcal{I}$,

number of candidates $n$, 

number of clusters $m$, 

similarity function $\texttt{sim}$, 

K-means clustering function $\texttt{KMeans}$, 

image captioning function $\texttt{Captioning}$

\State Initialize $\mathbf{S}_R \gets \{\}$
\While{$|\mathbf{S}_R| < n$}
    \State Append $\arg\max_{\mathbf{x} \in \mathcal{I}} \text{sim}(\mathbf{c},\mathbf{x})$ to $\mathbf{S}_R$
    \State Pop $\mathbf{x}$ from $\mathcal{I}$
\EndWhile
\State $\mathbf{S}_R^{(1)}, \ldots, \mathbf{S}_R^{(m)} \gets \texttt{KMeans}(\mathbf{S}_R)$
\State Define $\mathbf{p}_\mathbf{c}(\mathbf{x}) = \frac{\exp(\texttt{sim}(\mathbf{c},\mathbf{x}))}{\sum_{\mathbf{x}' \in \mathbf{S}_R} \exp(\texttt{sim}(\mathbf{c},\mathbf{x}'))}$
\State Initialize $\mathcal{T} \gets \{\}$
\For{$i = 1$ to $m$}
    \State $\hat{\mathbf{x}}^{(i)} \gets \arg\min_{\mathbf{x} \in\mathbf{S}_R^{(i)}}H(\mathbf{p}_\mathbf{x})$
    \State Append $\texttt{Captioning}(\hat{\mathbf{x}}^{(i)})$ to $\mathcal{T}$
\EndFor
\State \Return $\mathcal{T}$
\end{algorithmic}
\end{algorithm}
\begin{algorithm}[t]
\caption{Filtering Process}
\label{alg:filter}
\begin{algorithmic}[1]
\State \textbf{Input:} dialogue context $\mathbf{c}$, questions $\mathcal{Q}$,

retrieval candidates set $\mathbf{S}_R$,

similarity function $\texttt{sim}$, 

context answering function $\texttt{Answer}$

\State Define $\mathbf{p}_\mathbf{c}(\mathbf{x}) = \frac{\exp(\texttt{sim}(\mathbf{c},\mathbf{x}))}{\sum_{\mathbf{x}' \in \mathbf{S}_R} \exp(\texttt{sim}(\mathbf{c},\mathbf{x}'))}$
\State Define 

$\mathbf{p}_\mathbf{c,q}(\mathbf{x}) = \frac{\exp(\texttt{sim}(\texttt{concat}(\mathbf{c,q}),\mathbf{x}))}{\sum_{\mathbf{x}' \in \mathbf{S}_R} \exp(\texttt{sim}(\texttt{concat}(\mathbf{c,q}),\mathbf{x}'))}$
\State Initialize $\mathcal{Q}' \gets \{\}$
\For{$q$ in $\mathcal{Q}$}
    \If {\texttt{Answer}($\mathbf{c}, \mathbf{q}$) == ``uncertain''}
        \State Append $\mathbf{q}$ to $\mathcal{Q}'$
    \EndIf
\EndFor
\State $\hat{\mathbf{q}} \gets \arg\min_{\mathbf{q} \in \mathcal{Q}'}D_{KL}(\mathbf{p}_\mathbf{c}||\mathbf{p}_\mathbf{c, q})$

\State \Return $\hat{\mathbf{q}}$
\end{algorithmic}
\end{algorithm}

\paragraph{A plug-and-play approach.}
To address the issue of generated questions dealing with unrelated attributes to the target image, we inject the information about the retrieval candidate images of the current round as the textual input of the LLM questioner. For this process, we first extract images from the image pool that are similar to the (reformulated) dialogue context in the embedding space, establishing these as the set of ``retrieval candidates''. These similar images contain attributes analogous to the current dialogue context, which includes some information about the target image, ensuring that the questions generated about these attributes are somewhat guaranteed to be related to the target image. 

We apply K-means clustering to the candidate image embeddings. Subsequently, we obtain the similarity score distribution for each candidate image against the other candidates. For each cluster, the image with the lowest entropy in its similarity distribution is selected as the representative.
This selection is based on the rationale that a lower entropy in the similarity distribution
suggests that the corresponding image contains more concrete and distinguishable attributes. For example, among images belonging to the same cluster, the image corresponding to the caption ``\textit{home office}'' exhibits high entropy, while another image corresponding to the caption ``\textit{a desk with two computer monitors and a keyboard}'' exhibits low entropy.

The K images obtained through this method are then converted into textual information via an arbitrary image captioning model and provided as additional input to the LLM questioner. This \textit{retrieval context extraction} process is shown in Algorithm \ref{alg:extract}. To ensure the LLM questioner effectively grounds the textual information of the retrieval candidates, we utilize a chain-of-thought (CoT) approach. This includes providing the LLM questioner with few-shot examples as additional instructions, which involve the effective utilization of retrieval candidates. Appendix \ref{prompt} contains the CoT prompts provided to the LLMs.

Questions generated by grounding in the additional context extracted from the retrieval search space include attributes related to the target image but can still be redundant.
To prevent the generation of such questions, we employ an additional filtering process which is shown in recent work (\citealp{ddcot}).
For each question generated by the questioner, we prompt an LLM agent to respond with ``\textit{uncertain}'' if it cannot derive an answer from the corresponding description and dialogue, which implies the question is free from redundancy. We then only use questions answered with ``\textit{uncertain}''.

The filtering process can effectively remove questions answerable without viewing the target image but fails to exclude those that are unanswerable even with the target image present. 
These failure questions address attributes related to the candidates set but unrelated to the target image.
We observe that the use of such unsuitable questions causes a relatively abrupt change in the similarity distribution between the query and candidate images, resulting in decreased retrieval performance. Consequently, we select the question that exhibits the lowest Kullback-Leibler (KL) divergence of the similarity distributions about dialogue contexts and the distribution about dialogue contexts combined with the question. Algorithm \ref{alg:filter} shows the filtering process of PlugIR. The context-aware dialogue generation process, as configured in this manner, can be used synergistically with the context reformulation described in the previous sections. It also possesses the flexibility to be utilized independently, especially in scenarios where a fine-tuned retrieval model for the dialogue context is utilized.



\subsection{The Best log Rank Integral Metric}
\begin{figure*}[!t]
    \centering
    \begin{subfigure}[t]{.32\linewidth}
    \includegraphics[width=\linewidth]{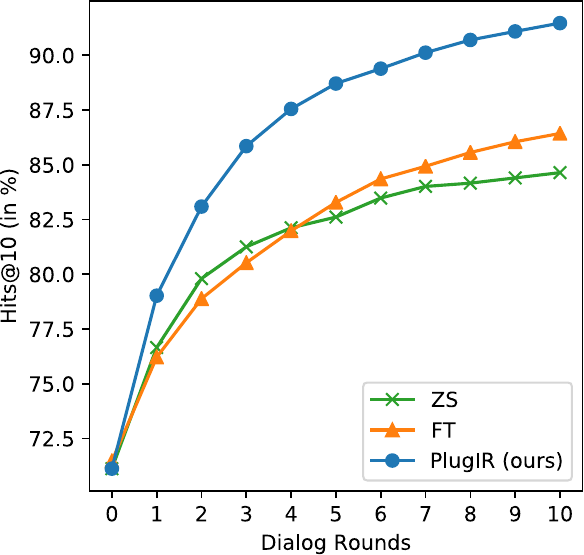} 
    \caption{VisDial}
    \label{fig:datasets1}
    \end{subfigure}    
    \begin{subfigure}[t]{.31\linewidth}
    \includegraphics[width=\linewidth]{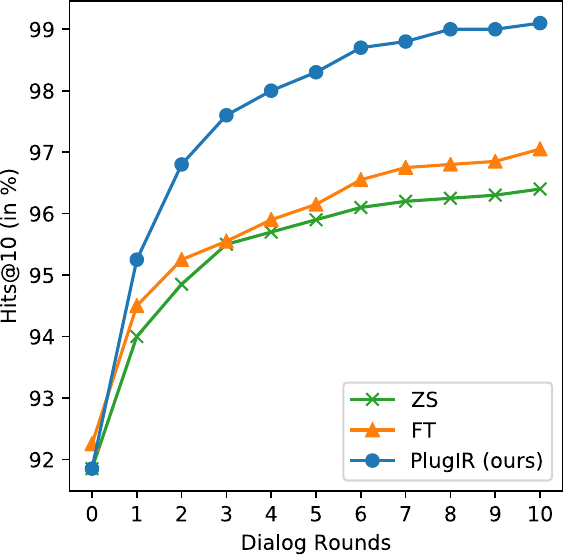} 
    \caption{COCO}
    \label{fig:datasets2}
    \end{subfigure}    
    \begin{subfigure}[t]{.31\linewidth}
    \includegraphics[width=\linewidth]{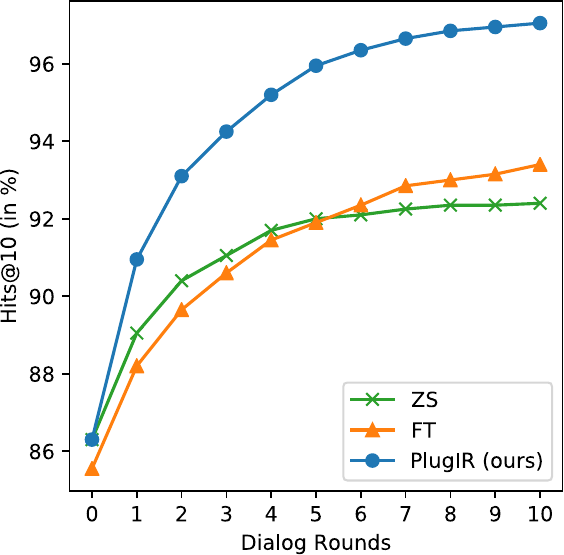} 
    \caption{Flickr30k}
    \label{fig:datasets3}
    \end{subfigure}
    \caption{
    Hits@10 comparisons of our proposed method with ZS and FT on VisDial, COCO, and Flickr30k.
    }\label{fig:datasets}
\end{figure*}
When evaluating an interactive retrieval system, the following key aspects are essential:
\begin{enumerate}[leftmargin=*, nolistsep]
    \item User satisfaction: This is considered fulfilled if the system manages to retrieve the target image at least once within its query budget.
    \item Efficiency: The system efficiency is gauged by rounds needed for successful retrieval; fewer rounds indicate better performance.
    \item Ranking improvement significance: Enhancements in higher ranking positions are intrinsically more challenging, and as such, they should be given more emphasis in metric evaluations. For instance, the improvement in metrics should be markedly more significant when an image's rank ascends from 2 to 1, as opposed to an ascent from 100 to 99. This distinction highlights the increased challenge and value associated with reaching the top rankings.
\end{enumerate}
Recall@K, commonly used for non-interactive retrieval system evaluation, falls short of fully addressing these three aspects in our specific context. Hits@K, recommended by \citet{levy2023chatting} for interactive systems, meets the criteria for user satisfaction but lacks in addressing the latter two aspects adequately. Consequently, this paper introduces a novel evaluation metric designed to comprehensively address all three of these considerations.

To address the aspect of user satisfaction, we define “\emph{Best Rank}” as follows:
\newtheorem{definition}{Definition}
\begin{definition} \label{def:best_rank}
Let $R(q)$ denote the retrieval rank of the target image corresponding to a query $q$. Then, the Best Rank $\pi$ for a query $q_t$ at round $t$ is
\[
\pi(q_t)=
\begin{cases}
\min(\pi(q_{t-1}), R(q_t))&\textup{if}\;t\geq1 \\
R(q_0)&\textup{if}\;t=0
\end{cases}.
\]
\end{definition}
\emph{Best Rank} measures the most successful retrieval out of all attempts up to each round. To reflect the second and third aspects, we introduce \emph{Best log Rank Integral} (BRI), defined using $\pi$ as follows:
\begin{definition}\label{def:bri}
    Let $Q$ and $T$ be a test query set and a designated system query budget, respectively. Then, BRI is defined as
    \[
    \mathop{\mathbb{E}}_{q\in Q}
    \left[
    \frac{1}{2T}\log\pi(q_0)\pi(q_T)+\frac{1}{T}\sum^{T-1}_{t=1}\log\pi(q_t)
    \right].
    \]
\end{definition}

BRI can be interpreted as the average area under the $\log\pi$ graph for round $t$ across all queries $Q$. The quicker the improvement in the ranks of target images, the less area there is under the graph. The logarithmic nature of the function causes a more substantial decrease in BRI as it nears the top ranks, with a lower BRI signifying better performance of the interactive retrieval system. Notably, BRI differs from Recall@K and Hits@K in its approach to evaluation. Rather than dichotomizing data samples based on a specific rank (K), it calibrates the results across all data samples for the evaluation, making it a more general and reliable metric.

To prove the reliability of BRI, we compare its correlation with human preference against those of previously proposed metrics in Section \ref{subsec:alignment}. The results confirm that \emph{BRI aligns considerably more closely with human evaluation than other metrics}.

\section{Experiments}

\subsection{Experimental Settings}

We evaluate our method on VisDial, COCO, and Flickr30k datasets. BLIP is used as the default text-to-image retrieval model unless explicitly stated, while BLIP-2 and ATM are also used for the experiment on adaptability. We employ ChatGPT \cite{ChatGPT} as the language model responsible for generating questions in all experiments, and BLIP-2 takes the place of human answerers in providing answers for the generated questions, considering the impracticality of human answerers.

In all experiments conducted in this paper, the number of clusters $m$ is uniformly set to $10$. Appendix \ref{appendix: clustering} presents the study about the effects of different $m$ values on the performance.

We report the results mainly with two metrics: Hits@10 and our proposed BRI. These evaluation metrics are selected because other metrics may lead to misinterpretations of interactive text-to-image retrieval systems' performance.  Examples substantiating this, along with further implementation details, are provided in Appendix \ref{experi_detail}.

\paragraph{Baselines and PlugIR.}
We compare our proposed PlugIR with two baselines, ZS and FT.
\begin{itemize}[leftmargin=*, nolistsep]
    \item{ZS}: A simple method that utilizes a zero-shot retrieval model. LLM questioner not aligned to retrieval context generates questions, and dialogues are directly used as queries for retrieval.
    \item{FT}: A method corresponding to \citet{levy2023chatting}, which uses a fine-tuned retrieval model. The rest are the same as ZS.
    \item{PlugIR}: Our proposed method that employs a zero-shot retrieval model. LLM questioner aligned to retrieval context generates questions, and revised dialogues via context reformulation are used as queries for retrieval.
\end{itemize}

\begin{table}[t]
    \centering
    \caption{Comparisons with baselines on VisDial, COCO, and Flickr30k}\label{table_results}
    \resizebox{1.0\linewidth}{!}{
    \begin{tabular*}{\columnwidth}{@{\extracolsep{\fill}}cccc@{}}
        \toprule
        \multirow{2}{*}{Method} & \multicolumn{3}{c}{BRI \(\downarrow\)} \\
        \cmidrule(lr){2-4}
        & VisDial & COCO & Flickr30k \\
        \midrule
        ZS & 1.0006 & 0.3576 & 0.5812 \\
        FT & 1.0106 & 0.3531 & 0.5793 \\
        PlugIR (ours)& \textbf{0.7674} & \textbf{0.2396} & \textbf{0.3733} \\
        \bottomrule
    \end{tabular*} 
}
\end{table}
\subsection{Results}
Table \ref{table_results} presents the BRI results of ZS, FT, and PlugIR on VisDial, COCO, and Flickr30k.
Across all three datasets, our proposed method outperforms both the
ZS and FT baselines.
Figure \ref{fig:datasets} demonstrates that PlugIR surpasses the baseline methods in all rounds in terms of the Hits@10 score.

A comparative analysis of the evaluation results for ZS and FT confirms that BRI provides a more comprehensive assessment compared to Hits@10. To be specific, in Figure \ref{fig:datasets1}, FT lags behind ZS until round 4, then outperforms ZS from round 5 onwards.
Consequently, at the end of the dialog rounds, Hits@10 indicates that FT outperforms ZS.
However, in the VisDial results shown in Table \ref{table_results}, we find that ZS and FT are comparable in terms of BRI. This is due to BRI considering not only the number of successful retrievals (user satisfaction) but also the number of rounds required for success (efficiency). Therefore, ZS's achievement of more successful retrievals in the early rounds results in a similar BRI score to FT.

\begin{table}[t]
    \centering
    \caption{Comparisons of BRI and existing metrics in terms of alignment with human preference}\label{table_human}
    \resizebox{1.0\linewidth}{!}{
    \addtolength{\tabcolsep}{-3.4pt}
    \begin{tabular*}{\columnwidth}{@{\extracolsep{\fill}}cccccc@{}}
        \toprule
        \multirow{2}{*}{Measure} & \multicolumn{5}{c}{Correlation coefficients with human} \\
        \cmidrule(lr){2-6}
        & Recall & MRR & NDCG  & Hits & BRI (ours) \\
        \midrule
        Spearman & 0.46 & 0.67 & 0.67 & 0.51 & \textbf{0.88} \\
        Pearson & 0.51 & 0.70 & 0.68 & 0.60 & \textbf{0.88} \\
        \bottomrule
    \end{tabular*} 
}
\end{table}

\subsection{BRI's Alignment with Human Evaluation}\label{subsec:alignment}
We engage 30 human testers to measure human preference for interactive text-to-image retrieval systems (for more details, please refer to Appendix \ref{sec:appen_human_eval}) and then explore its correlation with Recall, mean reciprocal rank (MRR), normalized discounted cumulative gain (NDCG), Hits, and BRI.
MRR and NDCG are metrics similar to Recall but additionally, consider the ranking improvement significance.
The correlations are quantified using the Spearman and Pearson correlation coefficients.
The findings presented in Table \ref{table_human} reveal that BRI is significantly more strongly correlated with human preference than the other metrics.

\begin{figure}[t]
{
\begin{center}
\centerline{\includegraphics[width=\columnwidth]{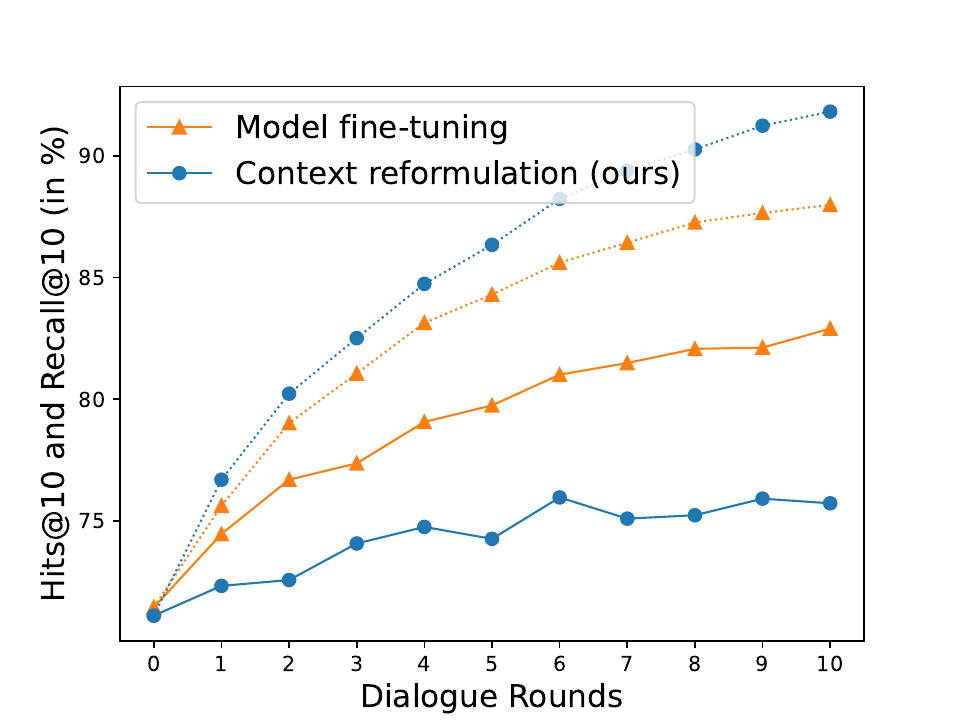}}
\caption{Round-by-round text-to-image retrieval performances of Model fine-tuning and context reformulation. Solid lines represent Recall@10, while dotted lines indicate Hits@10.}
\label{fig:dialog_utilization}
\end{center}
}
\end{figure}

\section{Analysis}
\subsection{Analyzing Dialogue Utilization in Model Fine-tuning and Context Reformulation}\label{sec:dialogues_utilization}
\begin{table}[t]
    \centering
    \caption{Experimental results across various retrievers} \label{adaptibility}
    \begin{tabular*}{\columnwidth}{@{\extracolsep{\fill}}cccc@{}}
        \toprule
        \multirow{2}{*}{Methods} & \multicolumn{3}{c}{BRI \(\downarrow\)} \\
        \cmidrule(lr){2-4}
        & BLIP & BLIP-2 & ATM \\
        \midrule
        ZS & 1.0006 & 0.8520 & 1.1329 \\
        PlugIR (ours) & \textbf{0.7674} & \textbf{0.6647} & \textbf{0.8236} \\
        \bottomrule
    \end{tabular*}
\end{table}

As previously illustrated in Figure \ref{fig:recon_motivationy}, zero-shot models have difficulties in understanding dialogues. In this section, we investigate whether model fine-tuning and context reformulation actually enhance the retrieval models' utilization of dialogues. We present the Recall@10 and Hits@10 for both model fine-tuning and context reformulation on the VisDial validation set in Figure \ref{fig:dialog_utilization}. The figure reveals that both methods, unlike the results in Figure \ref{fig:recon_motivationy}, achieve improved Recall@10 when augmenting queries with dialogues compared to using only image captions (round 0). Notably, we observe that these two methods behave differently in improving retrieval performance in our scenario. Model fine-tuning achieves a higher Recall@10 but a lower Hits@10 compared to context reformulation. This implies that model fine-tuning is more focused on succeeding in retrieval for the same samples that were successful in previous rounds, compared to context reformulation. Conversely, context reformulation, while less successful in retrieval per round compared to model fine-tuning, achieves a higher Hits@10 by improving dialogue utilization across the entire test query set.
In interactive text-to-image retrieval scenarios, the aggregate retrieval information accumulated up to each round holds more significance than the information from each round individually. The superior BRI of context reformulation compared to model fine-tuning reflects this aspect (see Table \ref{tab:robustness}).

\subsection{Adaptability to Various Pre-trained Models}
Due to its ability to function without the need for fine-tuning processes using dialogue datasets, PlugIR can utilize a wide range of retrievers, including black-box models. Table \ref{adaptibility} presents the results of evaluating PlugIR's performance using two additional retrievers, BLIP-2 and ATM, beyond the BLIP retriever used in previous experiments. It is observed that PlugIR outperforms the ZS baseline in all retriever settings, indicating that our methodology is not limited to specific retrievers and can be universally applied. Regarding performance across different retrievers, both ZS and PlugIR show similar trends, but a reduction in performance disparity between different retrievers is noted with PlugIR. This suggests that using PlugIR could reduce the cost associated with searching for an optimal retriever. Additional results for the Hits@10 metric can be found in Appendix \ref{appendix:addi_anal}.

\subsection{Robustness to Context Perturbation}
\begin{table}[t]
    \centering
    \caption{Robustness comparisons against context perturbations including character-level substitution and deletion (Char.) and style transfer to Informal, Slang, and Technical. $\Delta$ denotes the absolute performance degradation relative to Clean (no perturbation). We list the Hits@10 from the final (10th) round.}\label{tab:robustness}
    {\footnotesize
    \addtolength{\tabcolsep}{-2.1pt}
    \begin{tabular}{cccccc}
    \toprule
        Perturb. & Method & Hits@10 \(\uparrow\) & $\Delta$ & BRI \(\downarrow\) & $\Delta$\\ \midrule
        \multirow{3}{*}{Clean} & ZS & 84.98 & 0.00 & 1.0335 & 0.0000 \\ 
        & FT & 87.98 & 0.00 & 0.9987 & 0.0000 \\
        & PlugIR & \textbf{91.81} & 0.00 & \textbf{0.8507} & 0.0000 \\ \midrule
        \multirow{3}{*}{Char.} & ZS & 82.07 & 2.91 & 1.1255 & 0.0920 \\ 
        & FT & 84.54 & 3.44 & 1.1192 & 0.1205 \\
        & PlugIR & \textbf{91.04} & \textbf{0.77} & \textbf{0.8624} & \textbf{0.0117} \\ \midrule
        \multirow{3}{*}{Infor.} & ZS & 83.24 & 1.74 & 1.0721 & 0.0386 \\ 
        & FT & 87.06 & 0.92 & 1.0340 & 0.0353 \\
        & PlugIR & \textbf{90.94} & \textbf{0.87} & \textbf{0.8732} & \textbf{0.0225} \\ \midrule
        \multirow{3}{*}{Slang} & ZS & 82.75 & 2.23 & 1.0955 & 0.0620 \\ 
        & FT & 85.56 & 2.42 & 1.0780 & 0.0793 \\
        & PlugIR & \textbf{90.26} & \textbf{1.55} & \textbf{0.9082} & \textbf{0.0575} \\ \midrule
        \multirow{3}{*}{Tech.} & ZS & 81.69 & 3.29 & 1.1181 & 0.0846 \\ 
        & FT & 85.56 & 2.42 & 1.0701 & 0.0714\\
        & PlugIR & \textbf{89.78} & \textbf{2.03} & \textbf{0.9119} & \textbf{0.0612} \\
        \bottomrule
    \end{tabular}
\vskip -0.0in
}
\end{table}
In our scenario, users may have their unique speaking styles, leading to variations in the input distribution of the retrieval system. From the perspective of these context perturbations, we compare the robustness of our proposed method with those of ZS and FT. To ensure a fair comparison, we assume that each method has the same questioner and conduct experiments by perturbing user responses in a fixed dataset (VisDial validation set). The experimented context perturbations include character-level substitution and deletion, and style transfer \cite{styletransferllm} to "informal," "slang," and "technical" styles. We use the TextAttack library \cite{textattack} for character-level perturbations and GPT-3.5 for style transfer.
In Table \ref{tab:robustness}, we observe that our method employing LLMs to reformulate dialogues before using them as inputs to the retrieval model exhibits greater robustness against the tested context perturbations compared to directly using dialogues as inputs in ZS and FT models.

\subsection{Compatibility with Fine-tuned Models}

The context-aware dialogue generation (CDG) module that incorporates an LLM questioner also can be independently combined with the various retriever models, even with the models fine-tuned for texts with dialogue form.
Table \ref{table_comptibility} demonstrates the effective combination of the CDG with FT, indicating that our context-aware dialogue generation module can generate effective questions for various retrievers. 
Additional results about the Hits@10 metric can be found in Appendix \ref{appendix:addi_anal}.


\subsection{Ablation Study}

\begin{figure}[!t]
{
\begin{center}
\centerline{\includegraphics[width=\columnwidth]{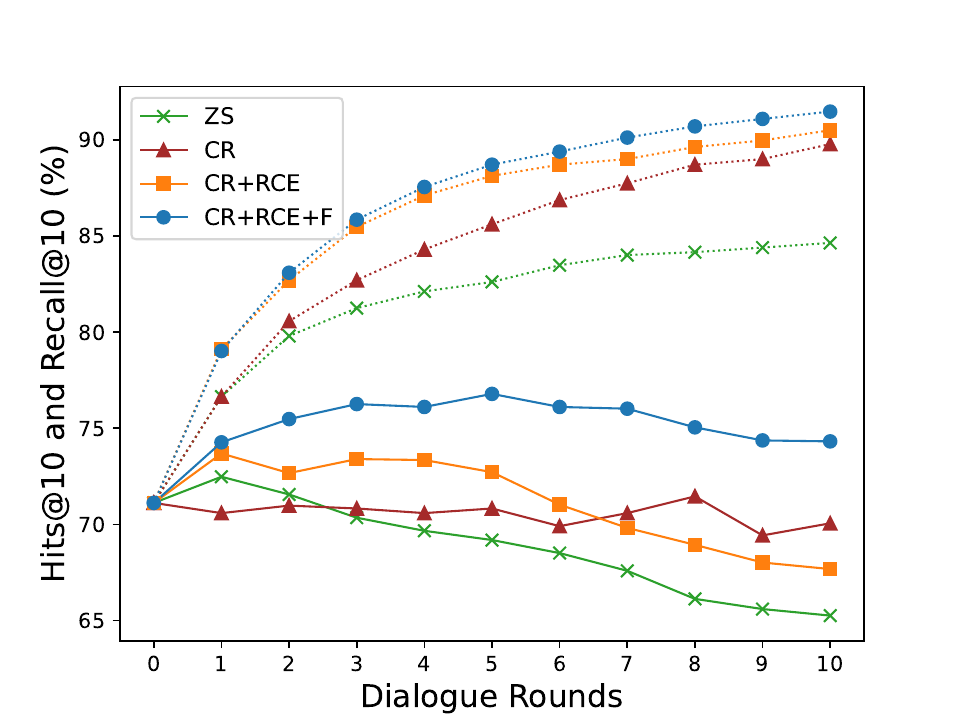}}
\caption{
Round-by-round text-to-image retrieval performances in the ablation study. Solid lines represent Recall@10, while dotted lines indicate Hits@10.
}
\label{fig:ablation}
\end{center}
}
\end{figure}

We conduct an ablation study on PlugIR, evaluating them in terms of Recall@10 and Hits@10. 
PlugIR comprises context reformulation (CR) and context-aware dialogue generation (CDG). The CDG can be subdivided into retrieval context extraction (RCE) and filtering (F) parts. Figure \ref{fig:ablation} presents the comparison of various combinations in terms of Recall@10 and Hits@10, which is conducted on the dialogue generated by the LLM questioner and answered by BLIP-2. Compared to the ZS baseline, each component progressively influences performance improvement in terms of Hits@10. Notably, the application of the F process leads to significant enhancements in Recall@10, which indicates, as mentioned in Section \ref{method}, an effective reduction in redundancy during question generation. 
The results of the ablation study for BRI performance are shown in Appendix \ref{appendix:addi_anal}. 

\begin{table}[t]
    \centering
    \caption{BRI results for the FT + CDG (Ours)}\label{table_comptibility}
    \begin{tabular}{lc}
        \toprule
        \multicolumn{1}{c}{Methods} & BRI \(\downarrow\) \\
        \midrule
        FT & 1.0106 \\
        FT + CDG & \textbf{0.9457} \\
        \bottomrule
    \end{tabular}
\end{table}

Additionally, further analysis in Appendix \ref{sec:app_further_analyses} demonstrates that PlugIR not only has a higher success rate in retrieval but also finds target images faster than FT. We provide a discussion about the hallucination of LLM agents of PlugIR in Appendix \ref{appendix: hallucination}.


\section{Conclusion}
We investigate the dialogue-form interaction with LLMs in the context of text-to-image retrieval.
Our proposed PlugIR progressively refines the text query for image retrieval using the dialogue between LLM questioner and the user.
Specifically, an LLM converts the dialogue into a format better understood by retrieval models.
PlugIR enables the direct application of an array of multimodal retrieval models, including black-box models, without necessitating further fine-tuning.
Moreover, we newly propose the Best log Rank Integral (BRI) metric allowing for the measurement of comprehensive performance in multi-turn tasks.
We verify the effectiveness of our retrieval system across a diverse range of environments, highlighting its versatile plug-and-play capabilities.

\section*{Limitations}
PlugIR adapts dialogues into a format compatible with pre-trained retrieval models. This means implementing PlugIR necessitates an understanding of the specific retrieval model. Our experiments show that the caption-style format is effective across all tested retrieval models and is likely to work with most others, although this is not universally guaranteed. Some retrieval models might benefit from alternative text query formats more than from the caption-style.
For example, the training datasets of recently proposed large vision language models \cite{llava, llava1_5} include various samples in dialogue format. Therefore, any interactive text-to-image retrieval system utilizing large vision-language models might prefer queries in dialogue form (we provide additional experiments in this direction in Appendix \ref{appendix:LVLM}). 
We identify this aspect as both a limitation of our current study and an avenue for future research.

\section*{Ethics Statement}
PlugIR demonstrates the capability to achieve effective performance in text-to-image retrieval tasks by leveraging the high-performance capabilities of a black-box multi-model text-image model and a large language model. However, this process poses a potential risk in retrieving individual data within the image pool. Additionally, there is a concern about the user's personal information leakage to the server operating the Large Language Model (LLM) during interactions between the LLM questioner and the user.

\section*{Acknowledgements}
This work was supported by the National Research Foundation of Korea (NRF) grants funded by the Korea government (Ministry of Science and ICT, MSIT) (2022R1A3B1077720), Institute of Information \& Communications Technology Planning \& Evaluation (IITP) grants funded by the Korea government (MSIT) (RS-2021-II211343: Artificial Intelligence
Graduate School Program (Seoul National University) and 2022-0-00959), Samsung Electronics Co., Ltd (IO221213-04119-01), and the BK21 FOUR
program of the Education and Research Program for Future ICT Pioneers, Seoul National University in 2024.


\bibliography{custom}

\clearpage
\appendix
\section{Prompting Examples for LLM} \label{prompt}

We show prompting examples for LLM in
Table \ref{table_prompt_baseline}, Table \ref{table_prompt_cot}, Table \ref{table_prompt_filter}, and Table \ref{table_prompt_recon}.
\section{Experimental Details} \label{experi_detail}
\subsection{Evaluation Metric}
In Sections \ref{sec:recon_motivation} and \ref{sec:dialogues_utilization}, we employed both Recall@10 and Hits@10 to assess the impacts of ZS, FT, and our proposed method. However, Recall@10 was not included in our main evaluations. This decision was based on the concern that using recall as an evaluation metric for interactive retrieval systems might create misleading perceptions of their effectiveness. We identified three key perspectives to be considered when evaluating these systems. Through illustrative examples, we demonstrate how conventional metrics can potentially misrepresent system performance, addressing each perspective in turn. For these examples, it is assumed that a user can view ten images per round.

\paragraph{User satisfaction.}

\begin{table}[t]
    \centering
    \caption{Retrieval results example using methods \textbf{A} and \textbf{B}, showcasing differences in the user satisfaction aspect}\label{app:table_user_satisfaction_data}
    \resizebox{1.0\linewidth}{!}{
    \begin{tabular*}{\columnwidth}{@{\extracolsep{\fill}}cccc@{}}
        \toprule
        \multirow{2}{*}{Method} & \multicolumn{3}{c}{Retrieval rank} \\
        \cmidrule(lr){2-4}
        & Round 0 & Round 1 & Round 2 \\
        \midrule
        \textbf{A} & 100 & 100 & 100 \\
        \textbf{B} & 100 & 10 & 100 \\
        \bottomrule
    \end{tabular*} 
}
\end{table}

\begin{table}[t]
    \centering
    \caption{Evaluation results of each metric for the two methods in Table \ref{app:table_user_satisfaction_data}}\label{app:table_user_satisfaction_results}
    \resizebox{1.0\linewidth}{!}{
    \begin{tabular*}{\columnwidth}{@{\extracolsep{\fill}}cccc@{}}
        \toprule
        Metric & \textbf{A} & \textbf{B} & Good or Bad \\
        \midrule
        Recall@10 \(\uparrow\) & 0.0 & 0.0 & Bad \\
        MRR@10 \(\uparrow\) & 0.0 & 0.0 & Bad \\
        NDCG@10 \(\uparrow\) & 0.0 & 0.0 & Bad \\
        Hits@10 \(\uparrow\) & 0.0 & 1.0 & Good \\
        BRI \(\downarrow\) & 4.6 & 2.9 & Good \\
        \bottomrule
    \end{tabular*} 
}
\end{table}

Table \ref{app:table_user_satisfaction_data} illustrates the retrieval ranks that methods \textbf{A} and \textbf{B} assigned to the same target image over three rounds. Table \ref{app:table_user_satisfaction_results} then shows the values of baseline metrics and BRI at the end of user-system interaction (in round 2). In Table \ref{app:table_user_satisfaction_data}, while the \textbf{A} method maintains a constant rank of 100 across three rounds, the \textbf{B} method allows the user to locate the target image in round 1, demonstrating its superior effectiveness compared to \textbf{A}. However, Table \ref{app:table_user_satisfaction_results} reveals that, except for Hits@10 and BRI, other metrics do not consider this aspect of interactive retrieval systems, resulting in both methods receiving the same evaluations.

\paragraph{Efficiency.}
\begin{table}[t]
    \centering
    \caption{Retrieval results example using methods \textbf{A} and \textbf{B}, showcasing differences in the efficiency aspect}\label{app:table_efficiency_data}
    \resizebox{1.0\linewidth}{!}{
    \begin{tabular*}{\columnwidth}{@{\extracolsep{\fill}}cccc@{}}
        \toprule
        \multirow{2}{*}{Method} & \multicolumn{3}{c}{Retrieval rank} \\
        \cmidrule(lr){2-4}
        & Round 0 & Round 1 & Round 2 \\
        \midrule
        \textbf{A} & 100 & 100 & 10 \\
        \textbf{B} & 100 & 10 & 10 \\
        \bottomrule
    \end{tabular*} 
}
\end{table}

\begin{table}[t]
    \centering
    \caption{Evaluation results of each metric for the two methods in Table \ref{app:table_efficiency_data}}\label{app:table_efficiency_results}
    \resizebox{1.0\linewidth}{!}{
    \begin{tabular*}{\columnwidth}{@{\extracolsep{\fill}}cccc@{}}
        \toprule
        Metric & \textbf{A} & \textbf{B} & Good or Bad \\
        \midrule
        Recall@10 \(\uparrow\) & 1.0 & 1.0 & Bad \\
        MRR@10 \(\uparrow\) & 0.1 & 0.1 & Bad \\
        NDCG@10 \(\uparrow\) & 0.3 & 0.3 & Bad \\
        Hits@10 \(\uparrow\) & 1.0 & 1.0 & Bad \\
        BRI \(\downarrow\) & 4.0 & 2.9 & Good \\
        \bottomrule
    \end{tabular*} 
}
\end{table}

In Table \ref{app:table_efficiency_data}, both methods \textbf{A} and \textbf{B} eventually assign a rank of 10 to the target image. However, the \textbf{B} method stands out because it enables the user to locate the target image in round 1, potentially reducing costs by obviating the need for round 2, thereby proving more efficient than the \textbf{A} method. Table \ref{app:table_efficiency_results}, however, reveals that all measures, except for BRI, overlook this efficiency aspect.

\paragraph{Ranking improvement significance.}

\begin{table}[t]
    \centering
    \caption{Retrieval results example using methods \textbf{A} and \textbf{B}, showcasing differences in the ranking improvement significance aspect}\label{app:table_ris_data}
    \resizebox{1.0\linewidth}{!}{
    \begin{tabular*}{\columnwidth}{@{\extracolsep{\fill}}ccc@{}}
        \toprule
        \multirow{2}{*}{Method} & \multicolumn{2}{c}{Retrieval rank} \\
        \cmidrule(lr){2-3}
        & Round 0 & Round 1 \\
        \midrule
        \textbf{A} & 100 & 10 \\
        \textbf{B} & 100 & 5 \\
        \bottomrule
    \end{tabular*} 
}
\end{table}

\begin{table}[t]
    \centering
    \caption{Evaluation results of each metric for the two methods in Table \ref{app:table_ris_data}}\label{app:table_ris_results}
    \resizebox{1.0\linewidth}{!}{
    \begin{tabular*}{\columnwidth}{@{\extracolsep{\fill}}cccc@{}}
        \toprule
        Metric & \textbf{A} & \textbf{B} & Good or Bad \\
        \midrule
        Recall@10 \(\uparrow\) & 1.0 & 1.0 & Bad \\
        MRR@10 \(\uparrow\) & 0.1 & 0.2 & Good \\
        NDCG@10 \(\uparrow\) & 0.3 & 0.4 & Good \\
        Hits@10 \(\uparrow\) & 1.0 & 1.0 & Bad \\
        BRI \(\downarrow\) & 3.5 & 3.1 & Good \\
        \bottomrule
    \end{tabular*} 
}
\end{table}

In Table \ref{app:table_ris_data}, although both methods \textbf{A} and \textbf{B} display the target image to the user in round 1, the \textbf{B} method assigns it a higher rank (nearer to 1) compared to \textbf{A}, making its retrieval more straightforward. Table \ref{app:table_ris_results} shows that metrics like MRR@10, NDCG@10, and BRI take this factor into account and consequently rate \textbf{B} more positively than \textbf{A}. However, Recall@10 and Hits@10 do not account for this particular aspect.

To summarize, employing traditional metrics like Recall@K in evaluating interactive retrieval systems can result in scores that only capture a fraction of these systems' capabilities. Basing system comparisons and assessments of superiority on such incomplete metrics might lead to misleading conclusions. For this reason, we opted not to use Recall@K as our main evaluation metric. Note that Hits@K has been utilized in prior research for assessing interactive retrieval systems, and we have included it in our main evaluation to facilitate comparisons.

Additionally, all scores except for BRI depend on the hyper-parameter of "the number of images a user can view at each round (@K)." However, this value can be highly variable, and thus the evaluation results can easily change. In contrast, BRI's independence from this hyper-parameter makes it a more stable and reliable evaluation metric.
Nevertheless, we provide additional analyses in Appendix \ref{sec:app_further_anal_conv_metrics}, using only conventional metrics including Recall, MRR, NDCG, and Hits.

\subsection{Further Implementation Details}
We utilze \texttt{gpt-3.5-turbo-0613} API as our LLMs. For hyperparameters, we use a temperature of 0.7 and a maximum token length of 32 for the question generation. For the context reformulation, we use a temperature of 0.0 and a maximum token length of 512. For filtering, we use a temperature of 0.0 and a maximum token length of 10. 

Regarding datasets, we generate dialogues for the entire 2,064 images in the VisDial validation set. Concerning COCO and Flickr30k, we generate dialogues on a sample of 2,000 images from each dataset. We set the number of candidates $n$ is set differently depending on the dataset used in the experiment; for VisDial, $n=500$, for COCO, $n=200$, for Flickr30k, $n=300$. This corresponds to approximately 1\% of the image pool (search space) size of each dataset.

\begin{figure}[!t]
{
\begin{center}
\centerline{\includegraphics[width=\columnwidth]{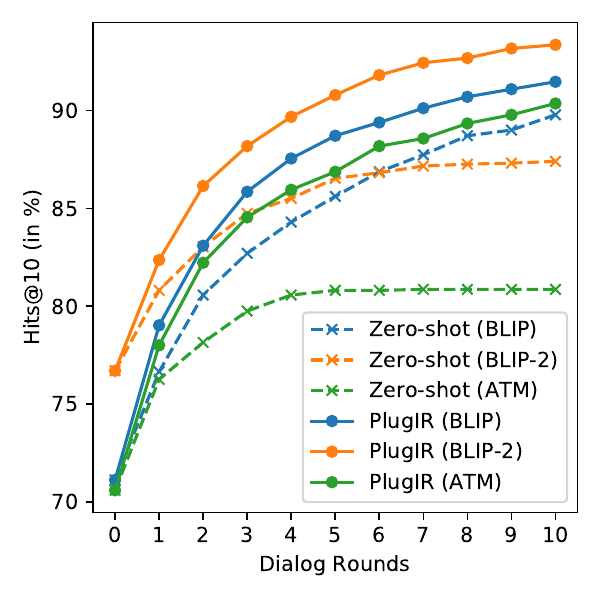}}
\caption{
Different retriever results.
}
\label{fig:retrievers}
\end{center}
}
\end{figure}
\section{Human Evaluation Details}\label{sec:appen_human_eval}
We measured human preferences of 30 machine-learning researchers who accepted to participate in the human evaluation as follows:
\begin{enumerate}[leftmargin=*, nolistsep]
    \item We prepared instances containing interactions between a user and an interactive text-to-image retrieval system. Each instance included {one target image, a 5-round dialogue, and the top-5 images for each round}. To facilitate the testers' evaluation (and to allow for easy recognition of the overall results at a glance), we decided to show five images for each of the five rounds for a single target image.
    \item 60 instances were presented to each of the 30 human testers, accompanied by the following instructions:
    
    ``Interactive image retrieval is a process where the system finds a target image desired by the user through interaction.
    When you evaluate, the scenario involves the user initially providing a brief description of the target image to the system. The system then formulates questions based on this description to correctly identify the target image, and the user responds to these questions.
    The process begins from round 0, where the user inputs a brief description into the system. From round 1 onwards, each round consists of the system asking a question about the target image based on the information exchanged with the user in previous rounds, and the user responding accordingly.
    At the end of each round, the system, based on the information gathered so far, searches for five candidate target images and presents them to the user in ranked order.
    The user can choose to end their interaction with the system at the conclusion of any round. Additionally, each round incurs monetary and time costs.
    You will evaluate the system's usefulness based on the interaction log on the left, which documents five rounds of interaction with the system.''
    \item For each instance, human testers are required to answer the following questions:
    
    ``Assuming you are using a system with the following log:
    
    Q1. The system will be effective.\newline(Yes: 5, No: 1)
    
    Q2. The system will be efficient.\newline(Yes: 5, No: 1)''
    
    Our questions aim to evaluate the system in terms of effectiveness and efficiency, which are generally considered in system assessments, without biasing the human testers' preferences.
    \item The average of the scores assigned to each instance is used as the human preference score.
\end{enumerate}
\begin{figure*}[!t]
    \centering
    \includegraphics[width=\textwidth]{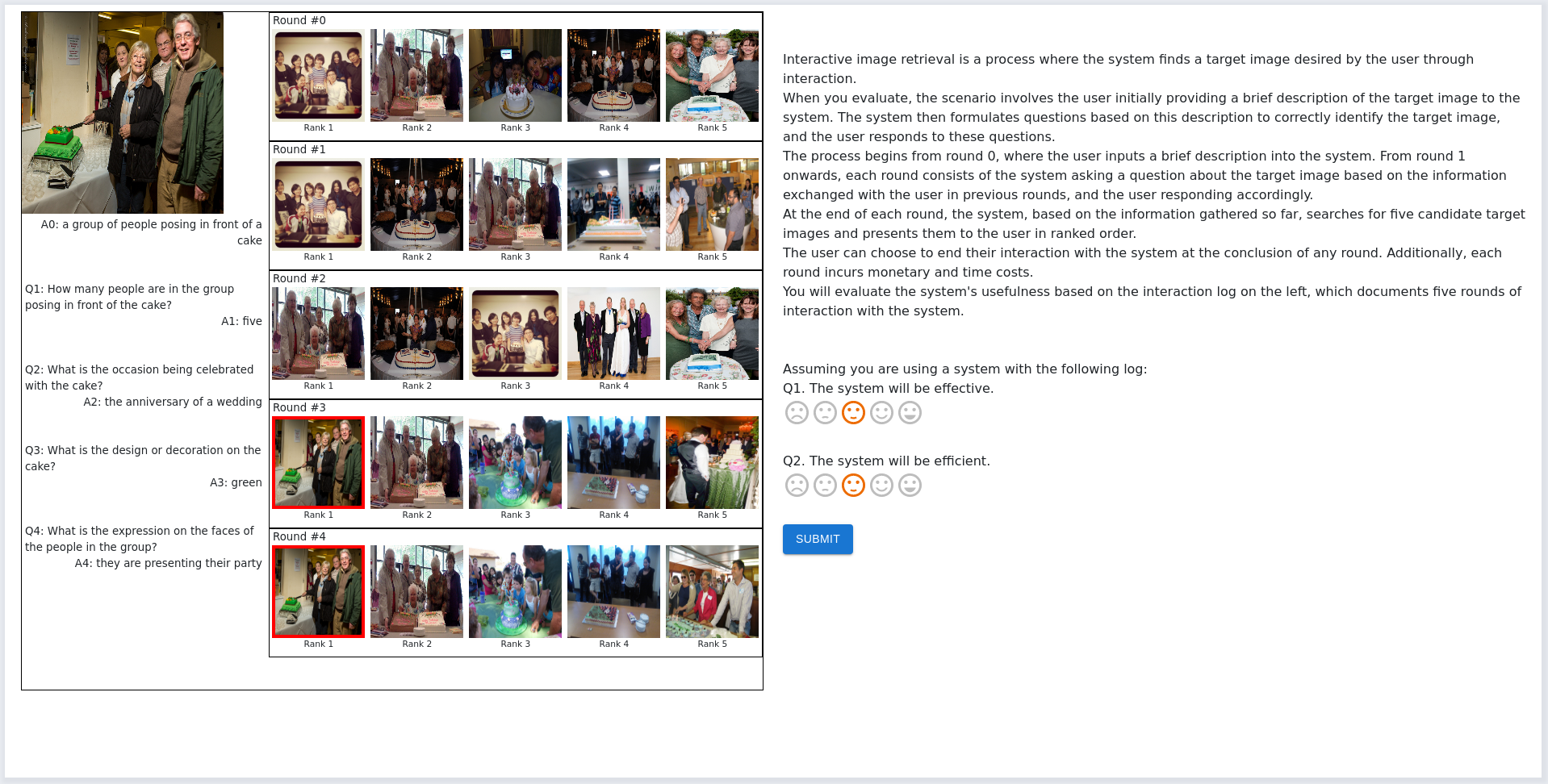} 
    \caption{
    Test screen used for measuring human preference
    }
    \label{fig:human_eval}
\end{figure*}
We provide a screenshot of the test in Figure \ref{fig:human_eval}.
\section{Additional Analyses Using Conventional Metrics}\label{sec:app_further_anal_conv_metrics}
We compare FT and our proposed method in a manner similar to Section \ref{sec:dialogues_utilization} of the manuscript. This comparison utilizes metrics such as Recall@10, MRR@10, NDCG@10, Hits@10, and the average number of rounds required for successful retrieval.

\begin{table*}[t]
    \centering
    \caption{Round-by-round text-to-image retrieval performances}\label{tab:app_total}
    \footnotesize\resizebox{\textwidth}{!}{
    \renewcommand{\arraystretch}{1.2}
    \begin{tabular}{c|c||ccccccccccc}
        \toprule
        \multirow{2}{*}{Metrics}&\multirow{2}{*}{Methods} & \multicolumn{11}{c}{Round} \\
        & & 0 & 1 & 2 & 3 & 4 & 5 & 6 & 7 & 8 & 9 & 10\\
        \midrule
        \multirow{2}{*}{Recall@10} & FT & 71.5  & 73.9    & 74.9    & 75.3   &76.0    &77.5    & 78.2   &78.3   & 78.8    &79.4   &79.5 \\
        & Ours & 71.1  &74.3  & 75.5 & 76.3  & 76.1&76.8 & 76.1 & 76.0 & 75.1 & 74.4 & 74.3    \\ \hline
        \multirow{2}{*}{MRR@10} & FT & 0.50   & 0.53    & 0.54     & 0.56   & 0.56   & 0.56   &0.57   &0.58   & 0.58  &0.58    &0.58\\
        &Ours & 0.51  & 0.53  & 0.56   & 0.56  & 0.57  & 0.57 & 0.56  &0.56  & 0.56 & 0.55&0.54   \\ \hline
        \multirow{2}{*}{NDCG@10} & FT & 0.55 & 0.58   & 0.59  & 0.60& 0.60   & 0.61 & 0.62  & 0.63& 0.63 & 0.63 &0.63 \\
        & Ours & 0.56  & 0.58 &  0.61  & 0.61 &  0.62  &  0.62   &  0.61    & 0.61  &  0.60   & 0.60    &  0.59    \\ \hline
        \multirow{2}{*}{Hits@10} & FT & 71.5  & 76.2   &78.9   &80.5 & 82.0   &83.3  & 84.4 &84.9  & 85.6&86.1  & 86.4 \\
        & Ours & 71.1& 79.0 & 83.1& 85.9& 87.6  & 88.7  & 89.4 & 90.1 &90.7 &91.1 & 91.5   \\
        \bottomrule
    \end{tabular} 
}
\end{table*}

MRR and NDCG enhance the concept of recall by incorporating the notion of ranking improvement significance. However, FT's higher MRR and NDCG scores than PlugIR's, as seen in the MRR@10 and NDCG@10 results of Table \ref{tab:app_total} beyond round 6, should not lead to the conclusion that FT outperforms PlugIR in terms of ranking improvement significance. This interpretation is cautioned against because a consistent trend is evident across the Recall@10, MRR@10, and NDCG@10 results. In other words, the superiority of FT's MRR and NDCG scores over PlugIR's is largely influenced by recall factors, rather than the ranking improvement significance. An in-depth analysis of these methods regarding recall is presented in Sections \ref{sec:cotext_reformulation} and \ref{sec:dialogues_utilization} of our manuscript as follows: FT achieves a higher Recall@10 but a lower Hits@10 compared to our proposed method. This implies that FT is more focused on succeeding in retrieval for the same samples that were successful in previous rounds, compared to ours. Conversely, the proposed method, while less successful in retrieval per round compared to FT, achieves a higher Hits@10 by improving dialogue utilization across the entire test query set. In interactive text-to-image retrieval scenarios, the aggregate retrieval information accumulated up to each round holds more significance than the information from each round individually. The superior BRI of PlugIR compared to FT reflects this aspect.

Nevertheless, we offer the following new analyses, not provided in our manuscript, by comparing the results of Recall@10, MRR@10, and NDCG@10 in Table \ref{tab:app_total}:

- In the results at the zeroth round of Table \ref{tab:app_total}, FT exhibits slightly higher recall than PlugIR but lower MRR and NDCG. This implies that fine-tuning the retrieval model on the visual dialogue dataset may disrupt the precise matching between captions and their corresponding images.

- In the round 4 and 5 results of Table \ref{tab:app_total}, while FT shows similar or slightly higher recall compared to PlugIR, it has lower MRR and NDCG. This indicates that PlugIR surpasses FT in terms of ranking improvement significance.
\section{Additional Analyses} \label{appendix:addi_anal}

\begin{figure}[t]
{
\begin{center}
\centerline{\includegraphics[width=\columnwidth]{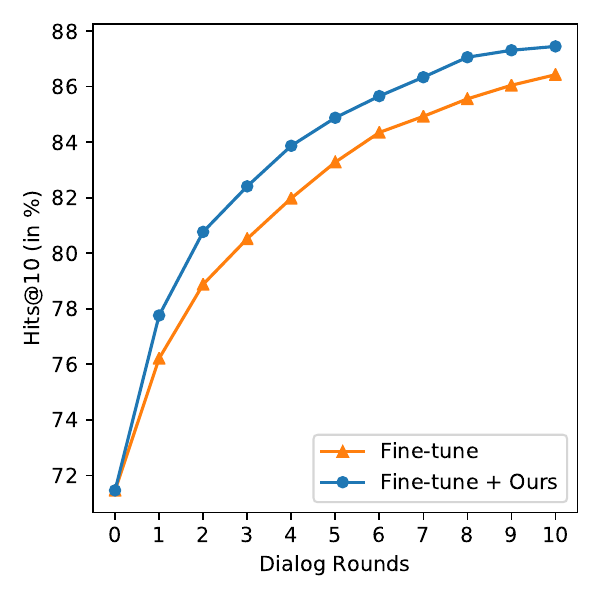}}
\caption{
Hits@10 results for the FT + CDG (Ours)
}
\label{fig:compatibility}
\end{center}
}
\end{figure}


\begin{table}[t]
    \centering
    \caption{Ablation study for BRI metric}\label{table_abl_bri}
    \begin{tabular}{lc}
        \toprule
        \multicolumn{1}{c}{Methods} & BRI \(\downarrow\) \\
        \midrule
        ZS & 1.0006 \\
        CR & 0.8907 \\
        CR + RCE & 0.7829 \\
        CR + RCE + F & \textbf{0.7674} \\
        \bottomrule
    \end{tabular}
\end{table}

We show Hits@10 results of analysis about the adaptability to various pre-trained models in Fig \ref{fig:retrievers}. Fig \ref{fig:compatibility} shows the Hits@10 results of analysis about the compatibility of the context-aware dialogue generation process with fine-tuned models. Table \ref{table_abl_bri} presents the BRI results of the ablation study.

\section{PlugIR Example} \label{example}

\begin{figure*}[!t]
    \centering
    \includegraphics[width=\textwidth]{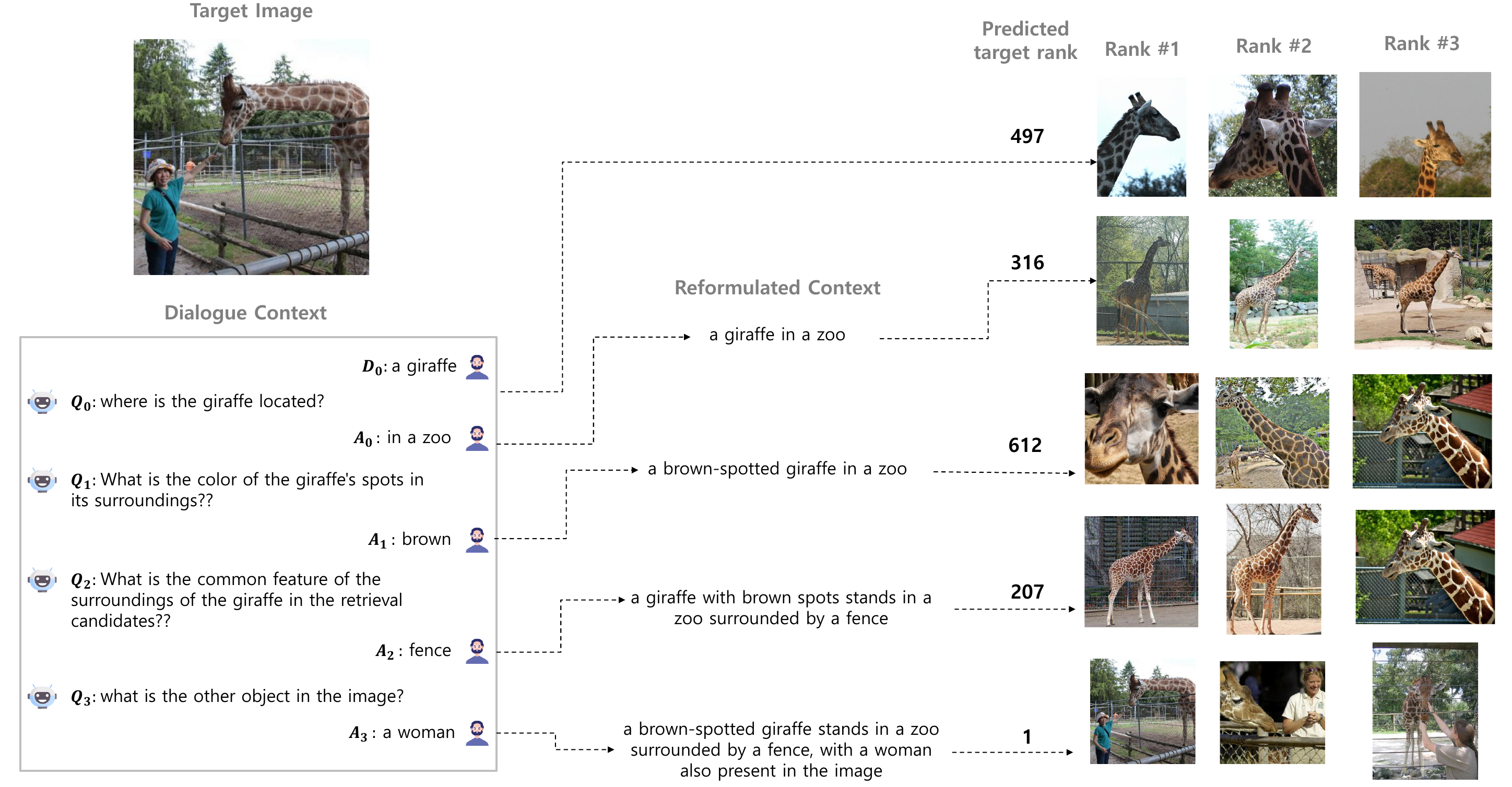} 
    \caption{
    The example of the plug-and-play interactive text-to-image retrieval system.
    }
    \label{fig:sample}
\end{figure*}

We show an example of PlugIR in Fig \ref{fig:sample}.
\section{Related Work} 
\label{related_work}

In this section, we explore potential baselines and state-of-the-art methods tailored for the image retrieval task, and discuss how existing methods differ from our approach in terms of methodology.

While the topic of interactive text-to-image retrieval is popular, existing methods in this field take a different approach with our method. Notably, \citet{guo2018dialog} and \citet{wu2021fashion} rely on one-sided feedback from the user, contrasting with our focus on a question-answer dialogue between the system and the user. Consequently, comparing performance of these lines of work with our method is challenging due to the different experimental setting. Instead, we provide a comparison from a perspective of methodology. \citet{guo2018dialog} and \citet{wu2021fashion} rely solely on user feedback without a questioner system, potentially leading to significant performance variations based on individual users. Moreover, users in their systems bear the burden of contemplating and providing feedback, contributing to substantial user fatigue. In contrast, our approach involves the system in actively formulating optimal questions for image retrieval. As a result, users only need to provide answers, significantly reducing user fatigue. Additionally, our method can minimize performance variations attributed to different users. This methodological distinction underscores the practical advantages of our setting, offering a more user-friendly and consistent interactive text-to-image search experience.

On the other hand, in the compositional image retrieval (CIR) field, there are recent papers related to interactive image retrieval. CIR methodologies typically involve the incorporation of not only textual information but also reference images to facilitate the retrieval of target images. Examples include \citet{baldrati2022effective} and \citet{karthik2023vision}. In particular, \citet{karthik2023vision} share a commonality with our approach by utilizing LLM. However, comparing performances with CIR methods poses a challenge due to the fundamental differences between two tasks, CIR and text-to-image retrieval, in that CIR additionally incorporates the use of reference images.

Upon thorough investigation, it appears that our approach and the baseline, ChatIR \cite{levy2023chatting}, are the only two approaches utilizing a question-answer dialogue in the current landscape of text-to-image retrieval. We believe that our focus on question-answer dialogue sets our method apart from existing methods, providing a distinctive angle in the realm of interactive text-to-image retrieval.


\section{Efficiency Comparison: Fine-Tuned vs. PlugIR}\label{sec:app_further_analyses}
\begin{table}[t]
    \centering
    \caption{The number of average rounds needed for successful retrieval (\#ARNSR)}\label{app:table_efficiency}
    \begin{tabular}{lc}
        \toprule
        \multicolumn{1}{c}{Methods} & \#ARNSR \\
        \midrule
        FT & 3.41 \\
        PlugIR (ours) & \textbf{2.85} \\
        \bottomrule
    \end{tabular}
\end{table}
We provide an analysis of the efficiency of PlugIR, our proposed method. While a high BRI score for PlugIR might not directly reflect its efficiency, as BRI encompasses a range of important factors beyond just efficiency, we turn our attention to a more intuitive measure of efficiency: the average number of rounds required for successful retrieval. Results from Table \ref{app:table_efficiency} indicate that PlugIR is more efficient than the tuning-based approach. Furthermore, the higher Hits@10 scores of PlugIR compared to those of FT indicate that our method not only finds target images faster than FT but also has a higher success rate in retrieval.





\section{Evaluating Query Style Preference in a Large Vision-Language Model: Caption versus Dialogue}\label{appendix:LVLM}

\begin{table}[t]
    \centering
    \caption{Preference of LLaVA-1.6-7B on dialogue-form vs. caption-form}\label{lvlm}
    \begin{tabular}{lcc}
        \toprule
        Rounds & Dialogue & Caption \\
        \midrule
        1 & 492 & 1572\\
        2 & 675&1389\\
        3&853&1211\\
        4&923&1141\\
        5&992&1072\\
        6&989&1075\\
        7&967&1097\\
        8&923&1141\\
        9&846&1218\\
        10&800&1264\\
        \bottomrule
    \end{tabular} 
\end{table}

we conducted the following experiment to investigate whether instruction-tuned large vision-language models also show a preference for caption- or dialogue-form:
\begin{enumerate}[leftmargin=*, nolistsep]
\item We create multiple-choice questions for each image in the VisDial validation dataset, each with one question and two choices.
\item The question is as follows: "What is more relevant to the photo?"
\item The first choice (Dialogue-form) is: "Caption: <caption>. Dialogue: <dialogue>." Here, "<caption>" is the image's caption, and <dialogue> is the dialogue from the VisDial sample.
\item The second choice (Caption-form) is: "Caption: <ours caption>." Here, "<ours caption>" is a caption created through our context reformulation of the first choice’s caption and dialogue.
\item To eliminate bias in the order of choices, we vary the order and feed the VQA sample to the model.
\end{enumerate}
We conducted the experiment on LLaVA-1.6-7B\footnote{https://huggingface.co/liuhaotian/llava-v1.6-vicuna-7b}, and the results are summarized in Table \ref{lvlm}. In Table \ref{lvlm}, "Rounds" indicates the length of the dialogue, with each round adding a set of question and answer. From Table \ref{lvlm}, we can see that the LLaVA model also prefers the caption form. Although there are intervals where the preferences appear to be similar (rounds 5-7), we can see that the preference for captions increases again as more information is provided.
\section{Hallucination Issues in Context-aware Dialogue Generation} \label{appendix: hallucination}

 In this section, we discuss the hallucination issues that appeared in the LLM agents in PlugIR, where LLMs are utilized within three steps of the pipeline: context reformulation, question generation, and filtering. We have occasionally discovered issues where parts of the content within a dialog are omitted during the context reformulation process. Additionally, during the filtering process, we have identified instances where, despite the LLM's capability to answer through dialog history, it classifies certain questions as non-redundant, deeming them unanswerable based on existing dialog content. Both of these instances can be linked to the hallucination problem in LLMs' inference processes, where some input content is ignored.

Moreover, we have observed a phenomenon where the format of questions generated by LLMs tends to conform to the structure of example questions provided in the prompt. For instance, if the example questions in the prompt predominantly inquire with "What," the questions generated by the LLM are mostly of the form such as "What is the color of the object in the photo?" This phenomenon becomes particularly pronounced when the explanations of how retrieval candidates were utilized in the added Chain of Thought (CoT) examples are specific. The more detailed the explanation, the more grounded the LLM questioner becomes in that example, enabling it to use retrieval candidates more effectively in generating questions. However, this also leads to the issue where the generated questions strongly adhere to the format of the example questions, due to the LLM questioner being heavily grounded not just in the explanation but also in the question format of the example. We believe this is related to the hallucination issue where LLM grounds on content considered noise within the given context, distracting the reasoning process.
\section{Effects of Different Clusters} \label{appendix: clustering}

 \begin{table}
     \centering
     \begin{tabular}{c|cccc}
     \toprule
        $m$  & 5  & 10  & 15  & 20 \\ \hline
         BRI \(\downarrow\) & 0.8742  & 0.8456 & 0.8280 & 0.8246\\
     \bottomrule
     \end{tabular}
     \caption{BRI scores about the various the number of clusters ($m$)}
     \label{table_clustering}
 \end{table}

Increasing the number of clusters ($m$) corresponds to increasing the number of captions injected into the LLM questioner, allowing the LLM questioner to refer to a more diverse set of characteristics from the retrieval candidates set. However, if the number of clusters increases beyond a certain level, captions sharing overlapping characteristics may become redundant and potentially hinder the LLM questioner's ability to generate correct questions. 

We conduct further studies on how the performance of BRI varies with different values of $m$. Table \ref{table_clustering} presents the results for PlugIR, which applies only context reformulation (CR) and retrieval context extraction (RCE) utilizing the 2024.02 version of ChatGPT. We observe that as the number of clusters increases, BRI performance improves. However, beyond a certain threshold, the extent of improvement becomes marginal ($m=15$ vs. $m=20$ comparison). This indicates that while increasing the number of clusters can initially contribute to enhancing BRI, there is a point of diminishing returns where further increases do not yield significant improvements. 

Similarly, setting the size of the retrieval candidate set ($n$) too small may fail to accurately grasp the context of the retrieval task, while setting it too large may increase the proportion of information unrelated to the target image in the image pool. Therefore, finding the optimal $m$ and $n$ for the image pool used by the user will also be a meaningful element in successfully applying our work. 
\section{Usage of AI Writing Assistance} \label{ai_Writing}

The paper was composed with linguistic assistance from AI assistant ChatGPT, which included paraphrasing, and spell-checking the author's original content.

\begin{table*}
\centering
{\small
\renewcommand{\arraystretch}{1.2}
\begin{tabular}{p{2cm} || p{13cm}}
\toprule
\textbf{System} 

\textbf{(Task} 

\textbf{Description)} & 
You are a proficient question generator tasked with aiding in the retrieval of a target image. Your role is to generate questions about the target image of the description via leveraging two key information sources:

[Description]: This is a concise explanation of the target image.
[Dialogue]: Comprising question and answer pairs that seek additional details about the target image.
Your generated question about the description must be clear, succinct, and concise, while differing from prior questions in the [Dialogue].
\\ \hline
\textbf{User} 

\textbf{(Train} 

\textbf{Example)} & 
[Description]
a man is doing a trick on a skateboard

[Dialogue]
Question: What type of trick is the man performing on the skateboard?
Answer: a jump
Question: What is the location of the jump trick being performed?
Answer: a skate park
Question: 
\\ \hline
\textbf{Assistant} 

\textbf{(Train} 

\textbf{Example)} & 
what is the outfit of the man performing the jump trick at a skate park?
\\ \hline
\textbf{User (Query)} & 
[Description]
\textbf{\{Initial Description\}}
[Dialogue]
\textbf{\{Dialogues\}}
Question:
\\
 \bottomrule
\end{tabular}
}
\caption{1-shot prompting example for LLM questioner of baseline.}
\label{table_prompt_baseline}
\end{table*}
\begin{table*}
\centering
{\small
\renewcommand{\arraystretch}{1.2}
\begin{tabular}{p{2cm} || p{13cm}}
\toprule
\textbf{System} 

\textbf{(Task} 

\textbf{Description)} & 
You are a proficient question generator tasked with aiding in the retrieval of a target image. Your role is to generate questions about the target image of the description via leveraging three key information sources:

[Retrieval Candidates]: These are captions of images which are the candidates of the retrieval task for the target image described in [Description].
[Description]: This is a concise explanation of the target image.
[Dialogue]: Comprising question and answer pairs that seek additional details about the target image.

You should craft a question that narrows down the options for the attributes of the target image through drawing the information from the retrieval candidates. The generated question about the target image must be clear, succinct, and concise. Also, the question should only be asked about common objects in the description and candidates, which cannot be answered only from the description and the dialogue. Please explain how did you utilize the information sources for generating a question.
\\ \hline
\textbf{User} 

\textbf{(Train} 

\textbf{Example)} & 
[Retrieval Candidates]
0. man in yellow shirt
1. a boy in a skateboard park
2. the biker is performing a trick
3. a man in a green hat doing half-pipe with a skateboard
4. a skateboarding man catches the air in the midst of a trick
[Description]
a man is doing a trick on a skateboard
[Dialogue]
Question: what type of trick is the man performing on the skateboard?
Answer: a jump
Question: what is the location of the jump trick being performed?
Answer: a skate park
Question:
\\ \hline
\textbf{Assistant} 

\textbf{(Train} 

\textbf{Example)} & 
what is the outfit of the man performing the jump trick at a skate park?

Explanation: To generate a question about the description, I will utilize the retrieval candidates that mention the outfit of the man. Candidates 0 and 3 provide information about the man's wearing. The description mentions the man's trick on a skateboard, and the dialogue mentions the type and the location of the trick. Since the attribute about the outfit does not appear in the description and the dialogue, the generated question cannot be answered from the information in the description and the dialogue about the target image. Also, the generated question is asking for the common objective, man, in the descriptions and candidates, not for the different objective from the description and the retrieval candidates 0 and 3, for example, a shirt and a half-pipe.
\\ \hline
\textbf{User (Query)} & 
[Retrieval Candidates]
\textbf{\{Retrieval Context\}}
[Description]
\textbf{\{Initial Description\}}
[Dialogue]
\textbf{\{Dialogues\}}
Question:
\\
 \bottomrule
\end{tabular}
}
\caption{1-shot prompting example for LLM questioner utilizing CoT and the additional context from the set of retrieval candidates.}
\label{table_prompt_cot}
\end{table*}
\begin{table*}
\centering
{\small
\renewcommand{\arraystretch}{1.2}
\begin{tabular}{p{2cm} || p{13cm}}
\toprule
\textbf{System} 

\textbf{(Task} 

\textbf{Description)} & 
Answer the question only according to the given context. If you cannot determine the answer or there are no objects that are asked by the question in the context , answer "Uncertain".
\\ \hline
\textbf{User (Query)} & 
[Context]
\textbf{\{Context\}}
[Question]
\textbf{\{Question\}}
[Answer]

\\
 \bottomrule
\end{tabular}
}
\caption{0-shot prompting example for LLM agent guided to answer the question according to the given context.}
\label{table_prompt_filter}
\end{table*}
\begin{table*}
\centering
{\small
\renewcommand{\arraystretch}{1.2}
\begin{tabular}{p{2cm} || p{13cm}}
\toprule
\textbf{System} 

\textbf{(Task} 

\textbf{Description)} & 
Your role is to reconstruct the [Caption] with the additional information given by following [Dialogue]. The reconstructed [New Caption] should be concise and in appropriate form to retrieve a target image from a pool of candidate images.
\\ \hline
\textbf{User} 

\textbf{(Train} 

\textbf{Example)} & 
[Caption]: a woman sits on a bench holding a guitar in her lap [Dialogue]: is this in a park? yes, i believe it is, are there others around? no, she is alone, does she have a collection bucket? no, is her hair long? yes, pretty long, is she wearing a dress? i don't think so, hard to tell, does she have shoes on? yes, flip flops, is there grass nearby? yes, everywhere, is it a sunny day? yes, are there trees? in the background there are trees, is the guitar new? i don't think so  [New Caption]:
\\ \hline
\textbf{Assistant} 

\textbf{(Train} 

\textbf{Example)} & 
a woman with pretty long hair sits alone on a grassy bench in a park on a sunny day, holding a guitar in her lap without a collection bucket, wearing flip flops, with trees in the background, with a slightly worn guitar
\\ \hline
\textbf{User (Query)} & 
[Caption]: \textbf{\{caption\}} [Dialogue]: \textbf{\{dialogue\}}  [New Caption]: 
\\
 \bottomrule
\end{tabular}
}
\caption{1-shot prompting example for LLM to reformulate the dialogue context.}
\label{table_prompt_recon}
\end{table*}

\end{document}